\documentclass[10pt, a4paper]{article}
\usepackage[utf8]{inputenc}

\usepackage[final]{lrec2026} 
\usepackage{booktabs}
\usepackage[table]{xcolor}
\usepackage{multirow}
\usepackage{tabularx}
\usepackage[most]{tcolorbox}
\usepackage{graphicx} 
\usepackage{longtable}
\usepackage{float}
\usepackage{caption}
\usepackage{booktabs, enumitem}
\usepackage[T1]{fontenc}
\usepackage{microtype}

\title{Rethinking Evaluation in Retrieval-Augmented Personalized Dialogue: A Cognitive and Linguistic Perspective}

\name{Tianyi Zhang, David Traum} 

\address{Institute for Creative Technologies, University of Southern California \\
         12015 Waterfront Drive, Los Angeles, CA 90094-2536, USA \\
         tzhang62@usc.edu, traum@ict.usc.edu\\}

\abstract{
In cognitive science and linguistic theory, dialogue is not seen as a chain of independent utterances but rather as a joint activity sustained by coherence, consistency, and shared understanding. However, many systems for open-domain and personalized dialogue use surface-level similarity metrics (e.g., BLEU, ROUGE, F1) as one of their main reporting measures, which fail to capture these deeper aspects of conversational quality. We re-examine a notable retrieval-augmented framework for personalized dialogue, LAPDOG, as a case study for evaluation methodology. Using both human and LLM-based judges, we identify limitations in current evaluation practices, including corrupted dialogue histories, contradictions between retrieved stories and persona, and incoherent response generation. Our results show that human and LLM judgments align closely but diverge from lexical similarity metrics, underscoring the need for cognitively grounded evaluation methods. Broadly, this work charts a path toward more reliable assessment frameworks for retrieval-augmented dialogue systems that better reflect the principles of natural human communication.
 \\ \newline  }

\begin{document}

\maketitleabstract

\section{Introduction}

Dialogue is not a sequence of isolated utterances; it is a collaborative process in which speakers build and maintain common ground~\citep{Clark-96,clark1991grounding}, manage discourse structure across turns~\citep{grosz1986attention}, and coordinate meaning through linguistic and conceptual alignment~\citep{pickering2004toward,poesio2001completions,mao2025chatgpt}. 
These requirements highlight why dialogue is a challenging task for natural language processing (NLP). Systems must preserve coherence over long and sometimes discontinuous histories, remain consistent with persona information, and avoid contradictions that would disrupt common ground~\citep{zhu2023pead}. One promising strategy for mitigating these issues is to give dialogue models access to external or supplementary information beyond the immediate conversation.  Retrieval-augmented generation (RAG) implements this idea by combining neural retrieval with response generation, allowing the model to dynamically access relevant background or persona-related content. 

We examine one recent approach to personalized dialogue generation, namely LAPDOG (Learning Retrieval Augmentation for Personalized Dialogue Generation) \citep{huang2023learning}, which augments persona profiles by retrieving additional external stories. We critically analyze and re-evaluate LAPDOG, considering issues with the history used, the evaluation metrics, and contradictions and coherence with both the dialogue history and external materials. 
We compare standard similarity metrics with judgments from both human annotators and LLM evaluators to assess whether current evaluation practices capture the cognitive and discourse-level qualities that make dialogue coherent and meaningful. Our results show that while similarity metrics report gains, human and LLM evaluators reveal inconsistencies—particularly in coherence and persona consistency—highlighting a disconnect between surface overlap and communicative quality.

Our main contributions are: \textbf{(1) Evaluation methodology:} We introduce a systematic framework that combines human and LLM judges, and analysis methods grounded in cognitive and linguistic theories of dialogue. This framework highlights key aspects of conversational quality—such as coherence, persona consistency, and engagement—that reflect how humans maintain common ground in interaction. \textbf{(2) Empirical findings:} We show that human and LLM judgments align closely but diverge sharply from surface metrics, revealing that improvements measured by BLEU, ROUGE, and F1 often fail to reflect gains in communicative or cognitive quality. This divergence highlights a deeper evaluative gap between linguistic form and interactive function, pointing to the need for metrics grounded in discourse and cognitive theory. \textbf{(3) Future directions:} We outline paths toward cognitively grounded evaluation, including taxonomies of persona–story relations and multi-objective training objectives that more directly model coherence, engagement, and the maintenance of common ground.

\begin{figure*}[t]
    \centering
    \includegraphics[width=\linewidth]{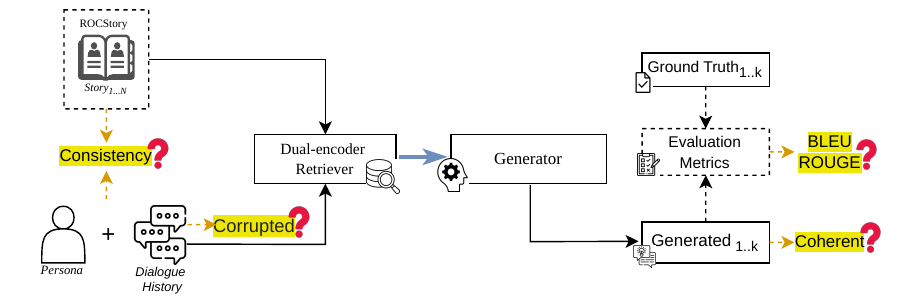}
    \caption{Overview of the LAPDOG retrieval-augmented personalized dialogue framework. The model retrieves external stories (e.g., from ROCStory) based on persona and dialogue history using a dual-encoder retriever, integrates them to a generator, and evaluates responses with metrics such as BLEU and ROUGE. The question marks indicate issues identified in our analysis—namely, use of similarity metrics, persona–story contradiction, incoherent generation, and dialogue corruption.}
    \label{fig:lapdog-overview}
\end{figure*}

\section{Related Work}
\label{sec:related_work}

Psycholinguistic theories of dialogue highlight how interlocutors update shared mental models and repair misunderstandings to preserve coherence~\citep{Clark-96,brennan1996conceptual}, while discourse theories emphasize the role of long-range dependencies and topic continuity in guiding interpretation~\citep{Grosz-95,hobbs1979coherence}. Cognitive models of alignment further suggest that successful dialogue depends on mutual adaptation at both linguistic and conceptual levels~\citep{garrod2009joint,pickering2013integrated}. Together, these perspectives show that dialogue is not only about producing locally appropriate sentences but also about maintaining consistency, coherence, and engagement across extended interactions.

\subsection{RAG for Personalized Dialogue}

Retrieval-Augmented Generation (RAG) has emerged as an effective framework for enhancing personalized dialogue by retrieving external content—such as documents, knowledge sentences, or narratives—and conditioning the generator on that content. These methods aim to go beyond predefined persona profiles by supplementing limited persona and dialogue history with richer context. By grounding responses in retrieved information, RAG can help maintain coherence over long dialogues, reinforce persona consistency, and reduce contradictions that stem from limited contextual awareness. 

Several recent RAG-based approaches, including UniMS-RAG \citep{li2024personalized} and SAFARI \citep{wang2023safari}, support multi-source integration by retrieving from persona memories, user profiles, and factual documents. PK-ICR \citep{oh2023pkicr} further improves retrieval by jointly selecting persona and knowledge pairs. However, these models generally assume access to sufficiently detailed persona stores and do not specifically target the problem of persona sparsity.

\subsection{LAPDOG}

LAPDOG (Learning Retrieval Augmentation for Personalized Dialogue Generation) \citep{huang2023learning} augments persona profiles by retrieving additional external stories. Retrieval candidates are selected using a dot-product similarity score between stories in a database and a persona-based dialogue query, and the retriever and generator are trained jointly with optimization based on similarity metrics. This approach improves performance on BLEU, ROUGE, and F1 compared to baseline models. However, these metrics emphasize surface-level lexical overlap and fail to capture deeper qualities of conversational competence such as coherence, engagement, and persona consistency \citep{liu-etal-2016-evaluate}.

LAPDOG \citep{huang2023learning} is distinctive in its use of narrative story data to enrich sparse personas through a fully end-to-end RAG framework. \citet{majumder2021story} also retrieve stories from the ROCstory dataset \citep{mostafazadeh2016corpus}, though their approach uses fixed similarity retrieval rather than a jointly optimized retriever–generator pipeline. More recent work has explored adaptive retrieval and discourse grounding for improved contextual coherence—e.g., \citet{ye2023retrieval} propose a retrieval–generation synergy model that dynamically updates persona representations during conversation, and \citet{zhan2024contextualrag} extend RAG with discourse-level contextualization to enhance turn-level consistency. Meanwhile, neuro-symbolic approaches have begun to bridge symbolic reasoning and neural retrieval for personalization; for instance, \citet{zhu2024neurosymbolic} integrate structured user representations with neural affective models to capture individual variation in sentiment and intent, while other recent systems adopt hierarchical or personality-guided architectures to ensure consistency and emotion alignment in the dialogue generation~\citep{zhu2024hippl, xie2025pgif}. 

LAPDOG's design illustrates the potential of retrieval augmentation. Its reliance on surface-level similarity metrics and certain structural limitations—such as corrupted histories, contradictions, and irrelevant retrievals—raise concerns that resonate with insights from cognitive science and linguistic theory. corrupted or fragmented dialogue histories prevent the model from maintaining the discourse structure that supports coherence across turns \citep{Grosz-95}. Similarly, when retrieved stories conflict with persona facts, the system breaks the common ground shared between speakers, which is essential for successful communication \citep{Clark-96}. Moreover, evaluation based only on lexical overlap neglects the cognitive and pragmatic dimensions of dialogue quality, such as engagement, consistency, and reasoning over shared knowledge.

\subsection{Personalized Dialogue Evaluation} \label{sec:related-Evaluation}

Many recent works in personalized or open-domain dialogue generation continue to report similarity metrics, such as BLEU, ROUGE, and F1 as their primary evaluation measures. For instance, \citet{Tang-23} evaluate contrastive latent-variable models for personalization largely through BLEU and ROUGE, while \citet{Lu-23} adopt similar metrics for assessing multi-attribute control in personalized dialogue. Even broader dialogue modeling efforts, such as \citet{Cheng-24} on in-dialogue learning, rely on these similarity-based scores as central indicators of quality. Although convenient, such metrics emphasize lexical overlap and provide little insight into discourse-level properties of conversation. From a cognitive science and linguistic theory perspective, this creates a misalignment: measures of surface similarity cannot capture coherence, consistency, or the maintenance of common ground that underlie natural dialogue \citep{Grosz-95,Clark-96}. 

Several studies have shown that n-gram overlap metrics (e.g., BLEU, ROUGE, METEOR) poorly reflect human judgments in dialogue evaluation. Liu et al.(2016) find these metrics fail to distinguish high-quality responses from random or baseline outputs \citep{liu-etal-2016-evaluate}, while Lowe et al.(2017) and Sharma et al.(2017) report similarly weak correlations in open-domain settings \citep{lowe2017ade, sharma2017diversity}. These results highlight the need for evaluation methods that better capture coherence, persona consistency, and contextual relevance in personalized dialogue.

Recent studies have shown that large language models (LLMs) can align well with human judgments across a variety of tasks \citep{chiang2023largelanguagemodelsalternative,wang2025dhpbenchmarkllmsgood}. However, this alignment is not universal. For instance, \citet{siro2024rethinking} found notable discrepancies between human and LLM evaluators in how user feedback was weighted during utterance evaluation. Similarly, \citet{reiss2023chatgpt} cautioned that ChatGPT’s utility in text classification is contingent on validation, due to task-specific inconsistencies. These findings stress the importance of validating LLM-based evaluation on a case-by-case basis. \cite{bavaresco-etal-2025-llms} survey many tasks and models and conclude that LLMs should be carefully validated against human judgments before being used as evaluators. To our knowledge, no prior work has established whether LLMs exhibit strong agreement with humans in both Likert ratings and relative rankings when evaluating persona-grounded dialogue responses. 

\section{Critical Analysis of Retrieval-Augmented Personalized Dialogue} \label{sec:crit-lapdog}

To better understand the challenges of retrieval-augmented approaches for personalized dialogue, we use LAPDOG as a representative case study. While LAPDOG demonstrates the promise of enriching persona profiles through external story retrieval, a closer look reveals several limitations, as illustrated in the system overview in Figure~\ref{fig:lapdog-overview}.
In the subsections that follow, we analyze four key aspects: the reliance on supervised similarity metrics,  incoherent response generation, contradictions between persona information and retrieved stories, and discontinuities caused by corrupted dialogue histories.

\subsection{Use of Supervised Similarity Metrics}
\label{subsec: simi-metrics}
LAPDOG was evaluated using F1, BLEU, and ROUGE-L to measure similarity to a reference response given the same context and persona. These metrics were also used during training and in the retrieval process for the generator. This framework implicitly assumes that the reference response represents the best possible outcome. While it may be reasonable to assume that a crowd-sourced human response is superior to most randomly generated machine outputs, it is not clear that lexical or even semantic similarity constitutes a meaningful dialogue evaluation function \cite{Liu-16}. Distinct responses can be equally or even more appropriate, while highly similar responses may be problematic due to a variety of pragmatic or contextual factors. Table~\ref{tab:similarity_failure} illustrates examples of a problematic response with high similarity metrics and an appropriate response with low similarity metrics.

\begin{table}[htbp]
\centering
\small
\renewcommand{\arraystretch}{1.2}
\begin{tabular}{p{2cm} p{5cm}}
\toprule
\textbf{Case} & \textbf{Example} \\
\midrule
\multirow{4}{=}{High Similarity\\Metrics} 
& \textit{Contradictory Response} \\
& \textbf{Reference:} Yes, I love going to Disney World every year. \\
& \textbf{Model:} No, I love going to Disney World every year. \\
& \textbf{BLEU / ROUGE-L / F1:} 89.32 / 88.89 / 88.89 \\
\midrule
\multirow{4}{=}{Low Similarity\\Metrics} 
& \textit{Appropriate Response} \\
& \textbf{Reference:} I enjoy visiting Disney World annually. \\
& \textbf{Model:} Yeah, we go there every year — it’s my favorite trip. \\
& \textbf{BLEU / ROUGE-L / F1:} 3.09 / 0.00 / 0.00 \\
\bottomrule
\end{tabular}
\caption{Illustrative limitations of surface-level similarity metrics. The first case shows high lexical overlap despite semantic contradiction. The second shows low lexical overlap despite pragmatic appropriateness.}
\label{tab:similarity_failure}
\end{table}

\begin{table}[htbp]
\centering
\small
\renewcommand{\arraystretch}{1}
\begin{tabular}{>{\bfseries}p{2.1cm} p{4.8cm}}  
\toprule
R Persona & 
I am a recluse. I am a famous twitch streamer. I am working on a building a robot to be my body double. I helped design the game starcraft. I've an allergy to water. \\
\midrule
Dialogue & 
{\bf R:} Hi. I'm Billy. I work as an electrician in Texas. \\
&{\bf Q:} Really? How? I do a lot of work on my mom and dads farm. \\
\midrule
LAPDOG (as R)& 
I'm a recluse and I build robots. I'm trying to be my body double. \\
\bottomrule
\end{tabular}
\caption{Abrupt introduction of persona content leading to incoherent response.}
\label{tab:abrupt_persona}
\end{table}

\subsection{Incoherent Response Generation} \label{subsec:incoherent}

Like other persona-driven dialogue systems, LAPDOG’s responses sometimes introduce information from the persona abruptly, resulting in a lack of contextual coherence with the preceding dialogue. For instance, as shown in Table~\ref{tab:abrupt_persona}, the LAPDOG system outputs the statement “I'm a recluse and I build robots. I'm trying to be my body double.” after a discussion of being an electrician in Texas. While this line reflects a relevant persona trait, its sudden appearance feels disconnected from the conversation's flow and is phrased awkwardly. This kind of jarring insertion can break the naturalness of the dialogue and leave hearers wondering how reclusiveness relates to electrical work. 
Such examples underscore the need for better integration strategies that can ground persona-related content in a conversational context more fluidly.
Better would be either showing relevance to the previous discussion \cite{sperber1986relevance} or explicitly signaling a proactive change in topic.

\begin{table}[htbp]
\centering
\small
\renewcommand{\arraystretch}{1}
\begin{tabular}{>{\bfseries}p{2.1cm} p{4.8cm}}
\toprule
Persona & 
I love to sing songs from the movie \textit{Frozen}. \textbf{I’m in the third grade.} I play with my friends on the playground. I love to go to Disney World every year. Mickey Mouse is my favorite character. \\
\midrule
Retrieved Story & 
I have always wanted to play a character at Disney World. 
 \textbf{Disney hired me as a customer service rep.} 
I moved to Orlando and applied for a job. 
I worked very hard to achieve my goal. 
 The other day I got a promotion to play Mickey Mouse. \\
\bottomrule
\end{tabular}
\caption{Inconsistency between persona and retrieved story content. Contradictory story lines are bolded.}
\label{tab:persona_conflict}
\end{table}

\subsection{Persona and Retrieved Story Contradictions} \label{subsec:contradictions}
Another significant issue involves the contradictions that arise between the external stories recovered and the established personas. For example, in Table~\ref{tab:persona_conflict}, the retrieved story does in fact share multiple features in common with the persona, such as Disney World and Mickey Mouse. But the persona indicates that the speaker is in the third grade, while the retrieved story describes an adult worker. This mismatch could lead to responses that break character, ultimately reducing the believability and personalization of the system. Such inconsistencies highlight the importance of retrieval filtering, relevance scoring, or more sophisticated integration mechanisms that can enforce alignment between persona and retrieved content.

\begin{table}[htbp]
\centering
\small
\renewcommand{\arraystretch}{1}
\begin{tabular}{r >{\raggedright\arraybackslash}p{6.3cm}}  
\toprule
\textbf{Line} & \textbf{Full Dialogue Context} \\
\midrule
1 & R: Cars are my thing. Vintage cars. I love working on them. Wrestling? Do you enjoy it? \\
\rowcolor{gray!15} 2 & Q: Yes, I love the crowds, getting to know people. \\
3 & R: I didn't think about the crowd aspect of wrestling. I do not like crowds. \\
\rowcolor{gray!15} 4 & Q: Understood. When i need to be alone, I work out a few times each week. \\
 5 & R: I agree. Working out is a great way to burn off steam. Do you like country music ? \\
\rowcolor{gray!15} 6 & Q: A little bit. I can get into taylor swift. \\
\rowcolor{gray!15} 7 & R: Lol. She's more pop now. Her old stuff was country. I like classic country. \\

\midrule
\textbf{Line} & \textbf{Corrupted Input Used by LAPDOG} \\
\midrule
2 & Q: Yes, I love the crowds, getting to know people.  \\
4 & R: Understood. When I need to be alone, I work out a few times each week .  \\
\rowcolor{gray!15} 6 & Q: A little bit. I can get into Taylor Swift .  \\
7 & R: Lol. She's more pop now. Her old stuff was country. I like classic country . \\
\bottomrule
\end{tabular}
\caption{Example original and corrupted dialogue.}
\label{tab:full_vs_corrupted}
\end{table}

\subsection{Discontinuous Dialogue History} \label{corruption-sec}
We identified that the original LAPDOG implementation removes and reassigns some utterances from the dialogue history during both training and inference. For example, Table~\ref{tab:full_vs_corrupted} illustrates a problematic training instance where LAPDOG constructs a new conversation using only utterances from line 2, 4, 6 and 7 of the original dialogue. This corruption introduces speaker inconsistencies and contextual disconnects. In the original dialogue, line 4 the utterance `` Understood. When I need to be alone, I work out a few times each week." 
is spoken by \textbf{Q}, but it is reassigned to \textbf{R} in the corrupted version, reversing the speaker roles. Additionally, line 6 the response ``A little bit. I can get into Taylor Swift.", 
originally served as a answer to respond the previous utterance asking: 
``Do you like country music?". Without this context, the dialogue loses coherence. 
Thus, corruption significantly disrupts conversational flow and context, negatively affecting dialogue coherence during both model training and evaluation. 
The evaluation using similarity to a reference utterance, described in section~\ref{sec:crit-lapdog} is even more problematic when the context for the utterance has changed, so that even the reference utterance may be incoherent.

\subsection{Discussion}
Taken together, the four issues discussed above raise doubts about whether LAPDOG is really an improvement over a baseline system that doesn't include retrieved stories. The first step is to re-evaluate LAPDOG, both on the full, uncorrupted dialogues, and using more appropriate evaluation metrics. We describe this in the next section. We plan to address the incoherence and contraditions issues in future work, using methods inspired by cognitive and linguistic theories of relevance and coherence rather than word-similarity.

\section{Evaluation Methodology} \label{sec:eval}


To assess both the evaluation metrics and LAPDOG’s reported gains, we conducted a comprehensive re-evaluation. We began by replicating the authors’ experiments using their official code, confirming that our reproduction matches the published numbers and preserves the same performance trend: LAPDOG consistently surpasses the baseline (see Table~\ref{tab:lapdog-repro}). To extend this analysis, we downloaded the full CONVAI2~\citep{convai2} conversations via ParlAI~\citep{miller2017parlai-lr} and re-trained both LAPDOG and the baseline model on this uncorrupted dataset. From this setup, we randomly selected 20 test dialogues and generated responses using the re-trained LAPDOG model, the baseline model, and the original ground-truth answers. The outputs were anonymized and shuffled before being evaluated by two human annotators—the second author and  a computer science master’s student—using the instructions shown in the Appendix A, 
 Table~\ref{tab:human-eval}.
In addition, two large language models (ChatGPT-o1 and DeepSeek-R1) were used to assess response quality with prompts like the example shown in Table~\ref{tab:llm-eval-prompt}. Both human and LLM evaluations employed two measures: a 1–5 Likert scale for overall quality (with 5 as best) and a ranking scheme (1 = best, 3 = worst) to capture relative preferences even when absolute scores were similar. For comparison, we also computed the similarity metrics used in~\citet{huang2023learning}. Finally, we computed average rating and ranking scores for both human and LLM annotations, along with an overall assessment combining all annotations.

\begin{table}
\centering
\footnotesize
\setlength{\tabcolsep}{2pt} 
\renewcommand{\arraystretch}{1} 
\begin{tabular}{lcccc}
\toprule
\textbf{Source(model)} & \textbf{System} & \textbf{F1} $\uparrow$ & \textbf{BLEU} $\uparrow$ & \textbf{ROUGE-L} $\uparrow$ \\
\midrule
Ours(T5-xl) & Baseline & 13.53 & 2.47 & 15.16 \\
             & LAPDOG   & \textbf{16.52} & \textbf{3.56} & \textbf{15.80} \\
\midrule
LAPDOG(T5-xl) & Baseline & 16.22 & 3.55 & 15.55 \\
                     & LAPDOG   & \textbf{17.11} & \textbf{3.56} & \textbf{15.64} \\
\bottomrule
\end{tabular}
\caption{Automatic similarity–metric scores from our reproduction (\emph{top}) and from the original LAPDOG paper (\emph{bottom}) using \textbf{corrupted} dialogue; higher is better for every metric. Bold marks the better system within each source block.}
\label{tab:lapdog-repro}
\end{table}

\begin{table}[t]
\centering
\small
\renewcommand{\arraystretch}{1}
\setlength{\tabcolsep}{6pt}
\begin{tabularx}{\linewidth}{X}
\toprule
\textbf{System Persona:} \\
\texttt{persona: i love to drink fancy tea. i have a big library at home. 
i'm a museum tour guide. i'm partly deaf.} \\
\midrule
\textbf{Dialogue Context:} \\
... \newline
\textbf{R:} i love reading. i have a big library at my house. \newline
\textbf{Q:} that's very nice, i love to read too and read about journals. \newline
\textbf{R:} i've a tough time hearing so reading is my happy place. \newline
\textbf{Q:} cool that is okay. my mom was very tough on me. \\
\midrule
\textbf{Candidate Responses:} \\
\textbf{R1:} \textit{"what do you do for a living?"} \newline
\textbf{R2:} \textit{"i'm sorry to hear that. i hope you have a wonderful life."} \newline
\textbf{R3:} \textit{"i work as a hearing impaired museum tour guide."} \\
\midrule
\textbf{Rating Prompt:} \\
Please rate each candidate response (R1, R2, R3) on a scale from 1 to 5 
based on overall response quality, 5 is the highest. \\
\midrule
\textbf{Ranking Prompt:} \\
Please rank each candidate response (R1, R2, R3) from 1 to 3, 
where \textbf{1 is your favorite response} and \textbf{3 is your least favorite}. \\
\bottomrule
\end{tabularx}
\caption{Example evaluation prompt used for LLM evaluations.}
\label{tab:llm-eval-prompt}
\end{table}

As noted in Section~\ref{sec:related_work}, alignment between LLM-based evaluations and human judgments is not universal. To assess this alignment in our setting, we applied two complementary analyses. First, we computed Pearson correlation coefficients between human annotators, LLM evaluators, and similarity metrics to quantify agreement across evaluation sources. Second, we used pairwise Williams tests \citep{williams1959comparison} to determine whether differences between correlations sharing a common variable were statistically significant. These analyses provide a principled basis for comparing the consistency of LLM and human judgments relative to traditional surface-level metrics.



\section{Empirical Findings} \label{sec:findings}
This section reports our empirical findings on retrieval-augmented personalized dialogue from both computational and cognitive perspectives. Grounded in theories of dialogue as a collaborative process requiring coherence, consistency, and shared understanding, we assess whether current model and metrics capture these properties. We first summarize quantitative results comparing LAPDOG, the baseline, and original responses across similarity metrics and human/LLM evaluations. We then examine the alignment between human and LLM judgments and their relation to surface-level similarity metrics, revealing the latter's limitations in capturing discourse-level and cognitive qualities.
\subsection{Main Results}
As established in Section~\ref{sec:eval} and Table~\ref{tab:lapdog-repro}, our reproduction of the LAPDOG setup matches the reported results and preserves the original performance trend. Building on this, Table~\ref{tab:comprehensive_evaluation} extends the evaluation to full, uncorrupted dialogues, enabling direct comparison with the corrupted setup. 

Despite overall improvement with full context, the baseline model outperforms LAPDOG on BLEU and ROUGE-L, while LAPDOG shows only a slight F1 gain. This suggests that LAPDOG’s previously reported advantage may stem from dataset or context differences rather than genuine modeling improvements. 

Human and LLM evaluations show a similar trend. As shown in Table~\ref{tab:comprehensive_evaluation}, both groups of evaluators consistently rate the original human responses highest, followed by the baseline, and then LAPDOG. Although the baseline receives slightly better average ratings and rankings than LAPDOG, the gaps are small and not statistically significant according to one-sided Wilcoxon~\citep{wilcoxon1945} signed-rank tests. The strong agreement between human and LLM evaluators indicates that they rely on similar criteria when judging response quality. These criteria may extend beyond surface similarity to include broader aspects of conversational adequacy. In the next subsection, we explore this possibility by examining how their evaluations relate to lexical similarity metrics.

\begin{table*}[ht]
\centering
\footnotesize
\setlength{\tabcolsep}{10pt}
\renewcommand{\arraystretch}{0.6}
\begin{tabular}{lccc||cc}
\toprule
\multicolumn{6}{c}{\textbf{Evaluation Results on Test Set}} \\
\midrule
\multirow{2}{*}{\textbf{Model}} & \multicolumn{3}{c||}{\textbf{Similarity Metrics}} & \multicolumn{2}{c}{\textbf{Overall Assessment}} \\
\cmidrule(lr){2-4} \cmidrule(lr){5-6}
& \textbf{F1} & \textbf{BLEU} & \textbf{ROUGE-L} & \textbf{Rating (1-5)} & \textbf{Ranking (1-3)} \\
\midrule
BASELINE T5$^S_{Sup}$(full) & 17.23 & \textbf{8.03} & \textbf{19.75}$^{**}$ & \textbf{2.71} & \textbf{2.16} \\
T5$^S_{Sup}$+LAPDOG(full) & \textbf{18.06} & 6.56 & 19.27 & 2.43 & 2.30 \\
Original Response & -- & -- & -- & 3.96 & 1.41 \\
\bottomrule
\end{tabular}
\vspace{2mm}
\begin{tabular}{lccccccc}
\toprule
\multicolumn{7}{c}{\textbf{Detailed Human and LLM Evaluation}} \\
\midrule
& \multicolumn{2}{c}{\textbf{Human Annotator 1}} & \multicolumn{2}{c}{\textbf{Human Annotator 2}} & \multicolumn{2}{c}{\textbf{Human Avg}} \\
\cmidrule(lr){2-3} \cmidrule(lr){4-5} \cmidrule(lr){6-7}
\textbf{Model} & \textbf{Rating} & \textbf{Ranking} & \textbf{Rating} & \textbf{Ranking} & \textbf{Rating} & \textbf{Ranking} \\
\midrule
BASELINE T5$^S_{Sup}$(full) & \textbf{2.60} & 2.30 & \textbf{3.05} & \textbf{2.15} & \textbf{2.83} & \textbf{2.23} \\
T5$^S_{Sup}$+LAPDOG(full) & 2.45 & \textbf{2.25} & 2.45 & 2.25 & 2.45 & 2.25 \\
Original Response & 3.60$^{**}$ & 1.45$^{**}$ & 4.10$^{**}$ & 1.25$^{**}$ &3.85$^{**}$ & 1.35$^{**}$  \\
\midrule
& \multicolumn{2}{c}{\textbf{o1}} & \multicolumn{2}{c}{\textbf{DeepSeek}} & \multicolumn{2}{c}{\textbf{LLM Avg}} \\
\cmidrule(lr){2-3} \cmidrule(lr){4-5} \cmidrule(lr){6-7}
BASELINE T5$^S_{Sup}$(full) & \textbf{2.55} & \textbf{2.05} & \textbf{2.65} & \textbf{2.15} & \textbf{2.60} & \textbf{2.10} \\
T5$^S_{Sup}$+LAPDOG(full) & 2.25 & 2.30 & 2.55 & 2.40 & 2.40 & 2.35 \\
Original Response & 3.95$^{**}$ & 1.55$^{*}$ & 4.20$^{**}$ & 1.40$^{**}$ & 4.08$^{**}$ & 1.48$^{**}$ \\
\bottomrule
\end{tabular}
\caption{Comparison of similarity metrics versus human/LLM evaluation. Top: similarity metrics and overall assessment. Bottom: detailed breakdown of individual evaluator scores. Rating: higher is better (1-5 scale); Ranking: lower is better (1 = best, 3 = worst). \textbf{Bold} numbers indicate better performance between baseline and LAPDOG models. Asterisks denote statistical significance of the difference between three responses
according to one-sided Wilcoxon signed-rank tests:
\,$^{*}\!p<0.05$, \,$^{**}\!p<0.01$.}
\label{tab:comprehensive_evaluation}
\end{table*}

\begin{table*}[t]
\centering
\footnotesize
\setlength{\tabcolsep}{6pt}
\definecolor{correlLight}{RGB}{242, 244, 248}
\definecolor{correlMedium}{RGB}{220, 226, 235}
\definecolor{correlDark}{RGB}{198, 208, 222}

\begin{tabular}{lcccccccc}
\toprule
& \multicolumn{2}{c}{\textbf{LAPDOG}} & \multicolumn{2}{c}{\textbf{Baseline}} & \multicolumn{2}{c}{\textbf{Original}} & \multicolumn{2}{c}{\textbf{Average}} \\
\cmidrule(lr){2-3} \cmidrule(lr){4-5} \cmidrule(lr){6-7} \cmidrule(lr){8-9}
\textbf{Annotator Pair} & \textbf{Rating} & \textbf{Ranking} & \textbf{Rating} & \textbf{Ranking} & \textbf{Rating} & \textbf{Ranking} & \textbf{Rating} & \textbf{Ranking} \\
\midrule
H1 \& H2 & \cellcolor{correlDark!76}0.600 & \cellcolor{correlDark!83}0.659 & \cellcolor{correlDark!62}0.489 & \cellcolor{correlDark!69}0.549 & \cellcolor{correlDark!28}0.222 & \cellcolor{correlDark!44}0.347 & \cellcolor{correlDark!55}0.437 & \cellcolor{correlDark!65}0.518 \\
H1 \& DS & \cellcolor{correlDark!92}0.734 & \cellcolor{correlDark!78}0.621 & \cellcolor{correlDark!34}0.269 & \cellcolor{correlDark!47}0.370 & \cellcolor{correlDark!44}0.348 & \cellcolor{correlDark!40}0.313 & \cellcolor{correlDark!57}0.450 & \cellcolor{correlDark!55}0.435 \\
H1 \& o1 & \cellcolor{correlDark!73}0.580 & \cellcolor{correlDark!58}0.459 & \cellcolor{correlDark!44}0.348 & \cellcolor{correlDark!40}0.315 & \cellcolor{correlDark!41}0.330 & \cellcolor{correlDark!12}0.096 & \cellcolor{correlDark!53}0.419 & \cellcolor{correlDark!36}0.290 \\
H2 \& DS & \cellcolor{correlDark!58}0.463 & \cellcolor{correlDark!67}0.533 & \cellcolor{correlDark!7}0.068 & \cellcolor{correlDark!47}0.374 & \cellcolor{correlDark!43}0.338 & \cellcolor{correlDark!64}0.508 & \cellcolor{correlDark!36}0.290 & \cellcolor{correlDark!59}0.471 \\
H2 \& o1 & \cellcolor{correlDark!91}0.723 & \cellcolor{correlDark!100}0.793 & \cellcolor{correlDark!40}0.313 & \cellcolor{correlDark!62}0.499 & \cellcolor{correlDark!75}0.597 & \cellcolor{correlDark!83}0.661 & \cellcolor{correlDark!69}0.544 & \cellcolor{correlDark!82}0.651 \\
DS \& o1 & \cellcolor{correlDark!100}0.791 & \cellcolor{correlDark!72}0.575 & \cellcolor{correlDark!14}0.143 & \cellcolor{correlDark!55}0.440 & \cellcolor{correlDark!72}0.570 & \cellcolor{correlDark!65}0.515 & \cellcolor{correlDark!63}0.501 & \cellcolor{correlDark!65}0.510 \\
\midrule
H1 \& BLEU  & \cellcolor{correlDark!1}0.006 & \cellcolor{correlDark!7}0.053 & \cellcolor{correlDark!36}0.287 & \cellcolor{correlDark!21}-0.166 & - & - & \cellcolor{correlDark!19}0.147 & \cellcolor{correlDark!7}-0.057 \\
H2 \& BLEU  & \cellcolor{correlDark!13}0.104 & \cellcolor{correlDark!27}-0.217 & \cellcolor{correlDark!9}-0.070 & 
0.037& - & - & \cellcolor{correlDark!2}0.017 & \cellcolor{correlDark!11}-0.090 \\
DS  \& BLEU & \cellcolor{correlDark!7}-0.053 & 
0.037 & 
\cellcolor{correlDark!11}0.086 & \cellcolor{correlDark!23}-0.180 & - & - & \cellcolor{correlDark!2}0.017 & \cellcolor{correlDark!9}-0.072 \\
o1  \& BLEU & \cellcolor{correlDark!21}0.165 & \cellcolor{correlDark!32}-0.253 & \cellcolor{correlDark!8}-0.058 & 
0.071 & - & - & \cellcolor{correlDark!7}0.054 & \cellcolor{correlDark!11}-0.091 \\
H1 \& ROUGE-L  & \cellcolor{correlDark!3}0.025 & \cellcolor{correlDark!18}-0.140 & \cellcolor{correlDark!5}-0.043 & 
0.563 & - & - & \cellcolor{correlDark!1}-0.009 & \cellcolor{correlDark!27}0.212 \\
H2 \& ROUGE-L  & \cellcolor{correlDark!50}0.397 & \cellcolor{correlDark!52}-0.412 & \cellcolor{correlDark!17}-0.133 & \cellcolor{correlDark!29}0.228 & - & - & \cellcolor{correlDark!17}0.132 & \cellcolor{correlDark!11}-0.092 \\
DS  \& ROUGE-L & \cellcolor{correlDark!3}0.027 & \cellcolor{correlDark!11}-0.092 & \cellcolor{correlDark!17}0.139 & \cellcolor{correlDark!14}-0.114 & - & - & \cellcolor{correlDark!10}0.083 & \cellcolor{correlDark!13}-0.103 \\
o1  \& ROUGE-L & \cellcolor{correlDark!24}0.188 & \cellcolor{correlDark!8}-0.066 & \cellcolor{correlDark!12}0.094 & \cellcolor{correlDark!7}0.056 & - & - & \cellcolor{correlDark!18}0.141 & \cellcolor{correlDark!1}-0.005 \\
H1 \& F1  & \cellcolor{correlDark!12}0.099 & \cellcolor{correlDark!16}-0.124 & \cellcolor{correlDark!11}0.088 & 0.479 & - & - & \cellcolor{correlDark!12}0.094 & \cellcolor{correlDark!22}0.178 \\
H2 \& F1  & \cellcolor{correlDark!51}0.407 & \cellcolor{correlDark!47}-0.372 & \cellcolor{correlDark!4}-0.030 & \cellcolor{correlDark!21}0.163 & - & - & \cellcolor{correlDark!24}0.189 & \cellcolor{correlDark!13}-0.105 \\
DS  \& F1 & \cellcolor{correlDark!6}0.044 & \cellcolor{correlDark!13}-0.103 & \cellcolor{correlDark!23}0.179 & \cellcolor{correlDark!19}-0.151 & - & - & \cellcolor{correlDark!14}0.112 & \cellcolor{correlDark!16}-0.127 \\
o1  \& F1 & \cellcolor{correlDark!23}0.182 & \cellcolor{correlDark!6}-0.044 & \cellcolor{correlDark!16}0.128 & \cellcolor{correlDark!3}0.023 & - & - & \cellcolor{correlDark!20}0.155 & \cellcolor{correlDark!1}-0.011 \\
\bottomrule

\end{tabular}
\caption{Pearson correlations among human annotators, LLM evaluators, and similarity metrics (BLEU, ROUGE-L, F1) across LAPDOG, baseline, and original answers. H1 and H2 refer to Human Annotators 1 and 2; DS is DeepSeek; o1 is a ChatGPT model, both LLM-based annotators. Darker shading indicates stronger correlation.}
\label{tab:correlations_combined}
\end{table*}

\subsection{Analysis of Human and LLM Judgments}

To further examine the relationship between human and LLM evaluations, we analyze their pairwise correlations across all measures. Table ~\ref{tab:correlations_combined} reports Pearson correlation between human ratings, LLM ratings, and automatic metrics. From a cognitive and linguistic perspective, this analysis allows us to assess whether LLMs approximate human interpretive processes--such as evaluating coherence, contextual fit, and persona consistency. 

The upper part of Table ~\ref{tab:correlations_combined} shows very high correlations for both ratings and rankings across human and LLM judges, with no clear distinction by evaluation categories. This indicates that human evaluators provide stable assessments of dialogue quality and that LLMs closely replicate these judgments.

Further statistical comparisons using the Williams test \citep{williams1959comparison} 
(Appendix B Table ~\ref{tab:williams-grid-final})
show no significant differences between human-LLM correlations ($p>0.05$), confirming that their evaluations are statistically indistinguishable. Overall, these findings demonstrate that LLM-based evaluators can reliably approximate human judgments of conversational quality.

\subsection{Limitations of Lexical Similarity Metrics}

Lexical similarity metrics such as BLEU, ROUGE, and F1 remain widely used in dialogue evaluation, yet they provide only a limited view of conversational quality. The following analysis compares these metrics with human and LLM judgments, revealing overlap-based measures emphasize lexical resemblance while overlooking deeper qualities such as coherence, persona consistency, and pragmatic appropriateness.

The lower part of Table~\ref{tab:correlations_combined} compares human/LLM judgments with BLEU, ROUGE-L, and F1. These correlations are substantially weaker and in some cases opposite to expectation (note that we would expect negative correlations with rankings), which indicates a clear divergence between lexical overlap metrics and human or LLM evaluations. To verify that this divergence is not due to random variation, we conducted Williams test (Appendix B Table~\ref{tab:williams-grid-final})
which confirms that these differences are statistically significant in nearly all cases ($p<0.05$). Specifically, correlations involving lexical overlap metrics such as BLEU, ROUGE, and F1 differ markedly from those based on human or LLM evaluations, demonstrating that overlap-based metrics capture a fundamentally different signal from the coherence and consistency that define effective dialogue.

From a cognitive and linguistic perspective, this divergence is not unexpected. Dialogue quality depends not only on surface resemblance but on the maintenance of coherence, persona consistency, and shared understanding—properties that emerge over multiple turns rather than in local n-gram overlap. Lexical metrics, which treat language as a sequence of tokens, cannot capture how interlocutors adapt to one another’s intentions, manage reference, or sustain common ground. Consequently, improvements reported under these metrics may not reflect genuine communicative competence.

Overall, the findings align with theories of dialogue that highlight coherence and common ground as key features of effective interaction. Future evaluation frameworks should therefore move beyond token-level overlap toward cognitively informed measures that capture the structural and pragmatic dimensions of natural conversation.

\section{Discussion and Future Directions}

In this paper, we provided a critical analysis of the LAPDOG pipeline, revealing key shortcomings in dialogue context continuity, persona consistency, and relevance.
Given the issues identified in Section~\ref{sec:crit-lapdog} and the re-evaluation in Section~\ref{sec:eval} and results in Section~\ref{sec:findings}, we propose several directions for future improvement in RAG-based personalized dialogue using stories. First, to address the limitations discussed in Section~\ref{subsec: simi-metrics}, we recommend replacing reference-based similarity metrics (e.g., BLEU, ROUGE) 
with LLM-based evaluators. As discussed in section~\ref{sec:related-Evaluation}, this is generally more appropriate for evaluation of dialogue capability, and as demonstrated in section~\ref{sec:findings}, there are high correlations with human judgments for evaluating persona-based dialogue.

To address the issues raised in Section~\ref{subsec:contradictions}, we propose improving the story selection process by incorporating a consistency classifier to filter out contradictory candidates. Contradictory information should be removed before stories are passed to the generator. We also propose categorizing candidate external sources based on their relationship to the persona: overlap, complementarity, and independence. Overlap refers to stories that restate persona facts and are already reflected in the profile—these offer minimal added value. Complementary stories enrich the persona with related but non-redundant information, helping expand depth without contradiction. Independent stories introduce new, unrelated topics and may add conversational breadth. We aim to study the impact of each category on coherence and personalization. 

To mitigate the coherence problems discussed in Section~\ref{subsec:incoherent},
we recommend incorporating metrics that analyze relevance and fluency, such as the FED fluency score FED \citep{mehri2020unsupervised}. We have addressed the issue in Section~\ref{corruption-sec} by using the full dialogue context from the original CONVAI2 dataset, and re-evaluated in Section~\ref{sec:eval}. Together, these contributions chart a clear path toward more coherent, persona-aware RAG dialogue systems that better integrate external story knowledge and align with human expectations.

Finally, it will be important to test the models in actual dialogue with the intended user population, rather than just offline analysis of individual dialogue turns.

\section{Conclusion}

This work re-evaluated retrieval-augmented personalized dialogue through a cognitive and linguistic lens, using LAPDOG as a representative case. We introduced a systematic evaluation framework that combines human and LLM judges with analysis methods grounded in theories of dialogue, emphasizing coherence, persona consistency, and engagement. Empirically, we found that human and LLM judgments align closely but diverge sharply from lexical similarity metrics, revealing a gap between linguistic form and communicative function. Building on these insights, we point to practical directions for system design. This moves retrieval-augmented dialogue systems toward behaviors that better reflect principles of natural human communication.

\section{Limitations}

Our work has several limitations. First, our hybrid human$+$LLM evaluation was tested using only two LLMs and two human raters on 20 example situations taken from the CONVAI dataset and augmented with stories from ROC-Story. It is unclear how well the results generalize to use of different resources. Second, our proposals still need to be fully implemented and validated to show improvements over LAPDOG and the baseline persona usage. Future work should therefore expand the evaluation and run additional experiments to confirm the reliability and usefulness of the proposed framework.


\section{Acknowledgements}
This work was supported by the U.S. Army Research Office under Cooperative Agreement Numbers W911NF-20-2-
0053 and W911NF-25-2-0040.


\section*{Bibliographical References}\label{sec:reference}

\bibliographystyle{lrec2026-natbib}
\bibliography{lrec2026-example}

\section*{Language Resource References}
\label{lr:ref}
\nocitelanguageresource{convai2}
\nocitelanguageresource{miller2017parlai-lr}
\bibliographystylelanguageresource{lrec2026-natbib}
\bibliographylanguageresource{languageresource}

\onecolumn
\section*{Appendices}

\subsection*{A Evaluation Instructions}

Table~\ref{tab:human-eval} gives the instructions and one dialogue example for human rating and rankings.
\vspace{1em}

\begin{longtable}{p{0.25\textwidth} p{0.7\textwidth}}
\toprule
\textbf{Section} & \textbf{Content / Evaluation Fields} \\ \midrule
\endfirsthead
\toprule
\textbf{Section} & \textbf{Content / Evaluation Fields} \\ \midrule
\endhead
\bottomrule
\\
\caption{Human Evaluation Sheet} 
\label{tab:human-eval} \\
\endfoot

\textbf{Task Definition} & 
For every test dialogue you will be shown the system's \textbf{persona}, the recent \textbf{dialogue context}, and three \textbf{candidate responses} (\texttt{R1}, \texttt{R2}, \texttt{R3}). 

Annotate each response based on:
\begin{enumerate}[leftmargin=1.2cm, nosep]
    \item \textbf{Absolute Quality (1--5):}
    \begin{itemize}[label={}, leftmargin=0.5cm, nosep]
        \item \textbf{5 Excellent:} Coherent, engaging, on-topic, persona-consistent.
        \item \textbf{4 Good:} Minor wording or relevance issues.
        \item \textbf{3 Adequate:} Bland or one noticeable flaw.
        \item \textbf{2 Poor:} Incoherent or ignores persona/context.
        \item \textbf{1 Unusable:} Nonsense, contradictory, or toxic.
    \end{itemize}
    \item \textbf{Relative Rank (1, 2, 3):} 1=best, 2=middle, 3=worst. Ties are allowed.
\end{enumerate} \\ \midrule

\textbf{System Persona} & \texttt{i love to drink fancy tea. i have a big library at home. i'm a museum tour guide. i'm partly deaf.} \\ \midrule

\textbf{Dialogue Context} & 

\textbf{Q:} hi how are you doing? i am okay how about you? \\
& \textbf{R:} i am doing good. just sipping tea. what do you do for work? \\
& \textbf{Q:} i used to do home health aide but now i am disabled. \\
& \textbf{R:} i am sorry to hear that. what happened \\
& \textbf{Q:} i graduated and got my license, worked a while, and became ill. \\
& \textbf{R:} so what do you do now for fun? i like to read. \\
& \textbf{Q:} i have dogs and i walk them. and a cat. \\
& \textbf{R:} what kind of dogs do you have? \\
& \textbf{Q:} i enjoy reading—read about michael jackson and psychology today. \\
& \textbf{R:} i love reading. i have a big library at my house. \\
& \textbf{Q:} that's very nice, i love to read too and read about journals. \\
& \textbf{R:} i've a tough time hearing so reading is my happy place. \\
& \textbf{Q:} cool that is okay. my mom was very tough on me. \\ \midrule

\textbf{Response 1 (R1)} & \textit{"what do you do for a living?"} \\
& \textbf{Quality (1--5):} \rule{1.5cm}{0.15mm} \quad \textbf{Rank (1--3):} \rule{1.5cm}{0.15mm} \\ \midrule

\textbf{Response 2 (R2)} & \textit{"i'm sorry to hear that. i hope you have a wonderful life."} \\
& \textbf{Quality (1--5):} \rule{1.5cm}{0.15mm} \quad \textbf{Rank (1--3):} \rule{1.5cm}{0.15mm} \\ \midrule

\textbf{Response 3 (R3)} & \textit{"i work as a hearing impaired museum tour guide."} \\
& \textbf{Quality (1--5):} \rule{1.5cm}{0.15mm} \quad \textbf{Rank (1--3):} \rule{1.5cm}{0.15mm} \\ \midrule

\textbf{Quality Legend} & 
\textbf{5:} Excellent (coherent, engaging, persona-consistent) \\
& \textbf{4:} Good (minor wording/relevance issues) \\
& \textbf{3:} Adequate (bland or one noticeable flaw) \\
& \textbf{2:} Poor (incoherent; ignores persona/context) \\
& \textbf{1:} Unusable (nonsense, contradictory, or toxic) \\
\end{longtable}

\newpage

\subsection*{B Correlation Difference Testing}

\noindent
\begin{minipage}{\textwidth}
\small
Table~\ref{tab:williams-grid-final} shows Williams tests results for correlations between variables, showing that human and LLM correlations are significantly higher than correlations with similarity metrics.
\vspace{1em}

\centering
\footnotesize
\setlength{\tabcolsep}{3pt}
\renewcommand{\arraystretch}{1.15}
\resizebox{\textwidth}{!}{
\begin{tabular}{cc}
\begin{minipage}[t]{.48\textwidth}\centering
\textbf{Shared variable: H1}

\begin{tabular}{@{}lrrrr@{}}
\toprule
Comparison & $r_1$ & $r_2$ & $t$ & $p$ \\
\midrule
H1--o1 vs.\ H1--BLEU   & 0.487 & 0.081 &  2.12 & 0.041$^{*}$\\
H1--DS vs.\ H1--BLEU   & 0.571 & 0.081 &  2.64 & 0.012$^{*}$\\
H1--H2 vs.\ H1--BLEU   & 0.555 & 0.081 &  2.48 & 0.018$^{*}$\\
H1--o1 vs.\ H1--ROUGE  & 0.487 & 0.038 &  2.33 & 0.026$^{*}$\\
H1--DS vs.\ H1--ROUGE  & 0.571 & 0.038 &  2.94 & 0.006$^{*}$\\
H1--H2 vs.\ H1--ROUGE  & 0.555 & 0.038 &  2.91 & 0.006$^{*}$\\
H1--o1 vs.\ H1--F1     & 0.487 & 0.073 &  2.15 & 0.038$^{*}$\\
H1--DS vs.\ H1--F1     & 0.571 & 0.073 &  2.81 & 0.008$^{*}$\\
H1--H2 vs.\ H1--F1     & 0.555 & 0.073 &  2.74 & 0.009$^{*}$\\
H1--H2 vs.\ H1--o1     & 0.555 & 0.487 &  0.67 & 0.510\\
H1--H2 vs.\ H1--DS     & 0.555 & 0.571 & -0.13 & 0.899\\
\bottomrule
\end{tabular}
\end{minipage}
&
\begin{minipage}[t]{.48\textwidth}\centering
\textbf{Shared variable: H2}

\begin{tabular}{@{}lrrrr@{}}
\toprule
Comparison & $r_1$ & $r_2$ & $t$ & $p$ \\
\midrule
H2--DS vs.\ H2--BLEU   & 0.309 & 0.015 & 1.37 & 0.179\\
H2--o1 vs.\ H2--BLEU   & 0.552 & 0.015 & 2.93 & 0.006$^{*}$\\
H2--DS vs.\ H2--ROUGE  & 0.309 & 0.149 & 0.77 & 0.447\\
H2--o1 vs.\ H2--ROUGE  & 0.552 & 0.149 & 2.21 & 0.034$^{*}$\\
H2--DS vs.\ H2--F1     & 0.309 & 0.165 & 0.71 & 0.481\\
H2--o1 vs.\ H2--F1     & 0.552 & 0.165 & 2.13 & 0.040$^{*}$\\
H2--H1 vs.\ H2--o1     & 0.555 & 0.552 & 0.03 & 0.974\\
H2--H1 vs.\ H2--DS     & 0.555 & 0.309 & 2.18 & 0.035$^{*}$\\
H2--DS vs.\ H2--o1     & 0.309 & 0.552 & -2.00 & 0.053\\
\bottomrule
\end{tabular}
\end{minipage}
\\[1.5em]
\begin{minipage}[t]{.48\textwidth}\centering
\textbf{Shared variable: o1}

\begin{tabular}{@{}lrrrr@{}}
\toprule
Comparison & $r_1$ & $r_2$ & $t$ & $p$ \\
\midrule
o1--DS vs.\ o1--BLEU   & 0.523 & 0.102 & 2.19 & 0.035$^{*}$\\
o1--DS vs.\ o1--ROUGE  & 0.523 & 0.092 & 2.30 & 0.027$^{*}$\\
o1--DS vs.\ o1--F1     & 0.523 & 0.091 & 2.35 & 0.024$^{*}$\\
o1--DS vs.\ o1--H1     & 0.523 & 0.487 & 0.34 & 0.733\\
o1--DS vs.\ o1--H2     & 0.523 & 0.552 & -0.22 & 0.826\\
\bottomrule
\end{tabular}
\end{minipage}
&
\begin{minipage}[t]{.48\textwidth}\centering
\textbf{Shared variable: DS}

\begin{tabular}{@{}lrrrr@{}}
\toprule
Comparison & $r_1$ & $r_2$ & $t$ & $p$ \\
\midrule
DS--o1 vs.\ DS--BLEU   & 0.523 & 0.050 & 2.52 & 0.016$^{*}$\\
DS--o1 vs.\ DS--ROUGE  & 0.523 & 0.095 & 2.28 & 0.029$^{*}$\\
DS--o1 vs.\ DS--F1     & 0.523 & 0.133 & 2.08 & 0.044$^{*}$\\
DS--o1 vs.\ DS--H1     & 0.523 & 0.571 & -0.45 & 0.652\\
DS--o1 vs.\ DS--H2     & 0.523 & 0.309 & 1.77 & 0.084\\
\bottomrule
\end{tabular}
\end{minipage}
\end{tabular}
}
\vspace{0.5em}
\captionof{table}{Williams tests for differences between two correlations that share one variable (40 paired dialogues). An asterisk ($^{*}$) indicates significance at $\alpha=0.05$ (two-tailed).}
\label{tab:williams-grid-final}
\end{minipage}

\end{document}